\def\year{2021}\relax
\documentclass[letterpaper]{article} % DO NOT CHANGE THIS
\usepackage{aaai21}  % DO NOT CHANGE THIS
\usepackage{times}  % DO NOT CHANGE THIS
\usepackage{helvet} % DO NOT CHANGE THIS
\usepackage{courier}  % DO NOT CHANGE THIS
\usepackage[hyphens]{url}  % DO NOT CHANGE THIS
\usepackage{graphicx} % DO NOT CHANGE THIS
\usepackage{amsmath}
\usepackage{natbib}  % DO NOT CHANGE THIS AND DO NOT ADD ANY OPTIONS TO IT
\usepackage{caption} % DO NOT CHANGE THIS AND DO NOT ADD ANY OPTIONS TO IT
\usepackage{accents}
\usepackage{placeins}
\usepackage[title]{appendix}
\usepackage[makeroom]{cancel}
\usepackage{multicol}
\usepackage{multirow}
\usepackage{verbatim}
\usepackage{latexsym}
\usepackage[linesnumbered,ruled,vlined,longend]{algorithm2e}
\usepackage{mathtools,amssymb,lipsum,nccmath}

\usepackage{cuted}
\usepackage{soul,color}
\title{Oscillatory Fourier Neural Network: A Compact and Efficient Architecture for Sequential Processing}
\author{
    Bing Han, Cheng Wang, and Kaushik Roy\\
}
\affiliations{
    School of Electrical and Computer Engineering, Purdue University\\
    {han183,wang4700,kaushik}@purdue.edu

}
\begin{document}

\maketitle

\begin{abstract}
Tremendous progress has been made in sequential processing with the recent advances in recurrent neural networks. However, recurrent architectures face the challenge of exploding/vanishing gradients during training, and require significant computational resources to execute back-propagation through time. Moreover, large models are typically needed for executing complex sequential tasks. % such as natural language processing. 
%% Since we have not used natural language processing datasets, is it a good idea to mention it here-done
To address these challenges, we propose a novel neuron model that has cosine activation with a time varying component for sequential processing. The proposed neuron provides an efficient building block for projecting sequential inputs into spectral domain, which helps to retain long-term dependencies with minimal extra model parameters and computation. 
%% A sentence on why the neuron model is good and easily trainable?-done
A new type of recurrent network architecture, named Oscillatory Fourier Neural Network, based on the proposed neuron is presented and applied to various types of sequential tasks. We demonstrate  that recurrent neural network with the proposed neuron model is mathematically equivalent to a simplified form of discrete Fourier transform applied onto periodical activation. In particular, the computationally intensive back-propagation through time in training is eliminated, leading to faster training while achieving the state of the art inference accuracy in a diverse group of sequential tasks. For instance, applying the proposed model to sentiment analysis on IMDB review dataset reaches 89.4\% test accuracy within 5 epochs, accompanied by over 35x reduction in the model size compared to LSTM.
The proposed novel RNN architecture is well poised for intelligent sequential processing in resource constrained hardware. 
% As an example IMDB sentiment analysis results can be highlighted here, like 10x fewer epochs to reach SOTA, 50% parameters needed...
%% Yes, we should add some results. -added IMDB results
\end{abstract}

\section{Introduction}
Recently, artificial neural networks, especially various types of recurrent neural networks (RNN) have demonstrated significant success in sequential processing tasks such as natural language processing (NLP)\cite{Yin2017}. 
However, complex sequential processing tasks desires increasingly large RNN models, and the training of RNNs encounters the notorious challenge of exploding or vanishing gradients when information across long time need to be preserved and processed. In particular, back-propagation through time (BPTT) applied on weight matrices associated with hidden states such as $W_{hh}$ need to unroll gradient calculations throughout large number of time steps. BPTT is not only prone to exponentially decaying or growing of the propagated values but also consuming significant computation resources over the recurrent steps. Although mitigating architectures, such as long-short term memory (LSTM) and gated recurrent units (GRU) have been proposed, these modified RNN models typically contain more training parameters, consumes longer time to train, and may still suffer the loss of gradients in presence of very long input sequences. 
On the other hand, it has been observed that extracting and understanding meaningful patterns at different time scales can be important for sequential processing tasks such as text sentiment analysis or body activity recognition. For instances, patterns in movie reviews could emerge at the word level, sentence/clause level, and paragraph level that altogether determine the outcome of the final sentiment of a review. In order to learn the features at various scales from sequential data, incorporating spectral analysis such as Fourier Transform in RNNs have received growing attention \cite{Lee2021}, \cite{prism_2020}. Typically, the input of temporal series can be fed into certain type of spectral processing such as Fourier transform, where frequency-domain outcome can be obtained and further analyzed by the subsequent neural network blocks.

In this work, we propose a new type of neuron with time varying cosine activation (termed TV-Cosine neuron), and construct an RNN architecture, named oscillatory Fourier neural network (O-FNN), for efficient learning for sequential tasks. Intuitively, the proposed neuron model projects sequential input data into a \textit{phase} of oscillating neuron, in contrast to conventional neurons such as ReLU or tanh that modulate the \textit{magnitude} of activation depending on the input. Moreover, each neuron in the proposed architecture is a cosine activation applied onto a superposition of input signal and time dependent rotating term of certain frequency.
The hidden state vector of the O-FNN accumulates the outcome of cosine activation across all time steps before being fed into the final layer of the network. Following this approach, no backpropagation through time is needed during the backward pass of training, and both forward and backward pass can process the sequential data in a fully parallel manner. We will show that after accumulating the cosine activation through time, the hidden states will equivalently represent a modified discrete Fourier transform of sine and cosine neural activation. 
%backprop through time not needed, backprop fully parallel in all timesteps
The major contribution is summarized as follows:
\begin{itemize}
    \item We propose \textbf{a new type of neuron with time varying cosine activation} for sequential processing. We demonstrate that RNN with such time-varying activation is mathematically equivalent to a simplified form of discrete Fourier transform. Due to the usage of cosine activation, the transform of data into frequency domain is more computationally efficient compared to applying Fourier transform on ReLU or sigmoidal neurons.
    \item We propose \textbf{a new type of RNN architecture, O-FNN}, that is fully parallelizable in both forward and backward passes. In particular, since the backward propagation is not unrolled in time,
    %% you have not mentioned the above earlier in the intro -done
    the issue of exponential explosion or decay of the gradient values across long range of time steps with conventional activation functions is eliminated.
    \item We show that O-FNN architecture is capable of achieving better performance on a plethora of sequential processing tasks with more compact models and high computational efficiency in comparison with regular RNN/LSTM. The characteristics of smaller memory footprint and faster training make the O-FNN especially suitable for deployment in resource-constrained hardware such as IOT and battery-powered edge devices.  
\end{itemize}

\section{Related Work}

While periodic activation functions have been proposed as early as 1980s \citep{Lapedes1987Nonlinear, mccaughan1997On}, learning with such periodic activation has received limited attention from the research community \citep{Sopena_1999, Wong_2002}. 
In \cite{Sopena_1999}, the authors show that with proper range of initial weight values, a multi-layer perceptron using sinusoidal neurons in the form of $\sin{(WX+b)}$  improves accuracy and trains faster compared to its sigmoidal counterpart on some small datasets.  
% In addition, it is found that the improvement of sine neurons over sigmoidal neurons is sensitive to the range of initial weights. In \ref{Wong_2002}, improvement is achieved in a hand-written digit classification task when a hybrid periodical activation function is used instead of monotonic activation. The improvement is attributed to the stronger representation capability associated with the non-linearity in the periodic activation. 
Recently, sinusoidal activation function in implicit neural representations demonstrated improved capability of processing complex spatial-temporal signals and their derivative  \cite{sitzmann_2020}. Authors in \cite{search_2017} conduct network architecture search and identify neuron with partial sinusoidal behavior as one of the top candidates for activation functions for typical image classification tasks. Note that training networks with periodical activation may be challenging due to the possibility of having numerous local minima in the landscape of loss function \cite{Sopena_1999, taming_wave}.
%Meanwhile, in comparison with the predominant usage of monotonic functions, implementing a sinusoidal type of activation function is largely hindered by the paramount challenge of having multiple local minima in the loss function landscape during training. In \ref{taming_wave}, the authors formally demonstrate the occurrence of local minima encountered when training a multiple layer perceptron with sinusoidal neurons. 
%It shows that in the case of successfully trained networks, the periodicity of the sinusoids is often largely unused, since the input range of sine function after training is constrained close to 0, and sin(x) behaves similarly to traditional neuron tanh(x) in proximity of x=0. 

% \item RNNs with sinusoidal activation
Regarding applying periodical activation in RNNs, \cite{sopena_1994} reports improvement in learning when sine instead of sigmoid activation function is used in the last fully connected layer of a simple RNN trained for a next-character  prediction task. The authors of \cite{taming_wave} also observe that the periodical activation can be beneficial for training RNNs and LSTMs on some algorithmic tasks, showing faster learning and possibly better accuracy. Moreover, oscillatory neuron dynamics is exploited in implementing coupled oscillator recurrent neural networks in \cite{rusch2021coupled}, demonstrating great potential of oscillating neurons for processing complex sequential data.

%\item Fourier type of activation google, the Halawa[2008] paper
Usage of periodical activation is related to Fourier transforms which also involve extraction of information and features in the frequency domains. In particular, Fourier transforms, especially discrete Fourier Transforms (DFT), have been adopted successfully for RNNs to execute sequential processing \cite{koplon_1997} or make predictions \cite{forenet_2000}. Fourier transforms are typically done on input signals to facilitate the learning of spectral features. 
More recently, the idea of leveraging Fourier transforms for extracting spectral information is also explored in Natural Language Processing (NLP) tasks. Authors in \cite{FRU_2018} proposes using Fourier basis to summarize the statistics of hidden states through past time steps in RNNs.  \cite{prism_2020} applies spectral filters similar to DFT to the activations of individual neurons in BERT \cite{bert}
%% reference?
language model, aiming for extracting information changing at different time scales in texts.

\section{Proposed Approach}

\begin{figure}[ht!] 
\centering
\includegraphics[width=3.3in]{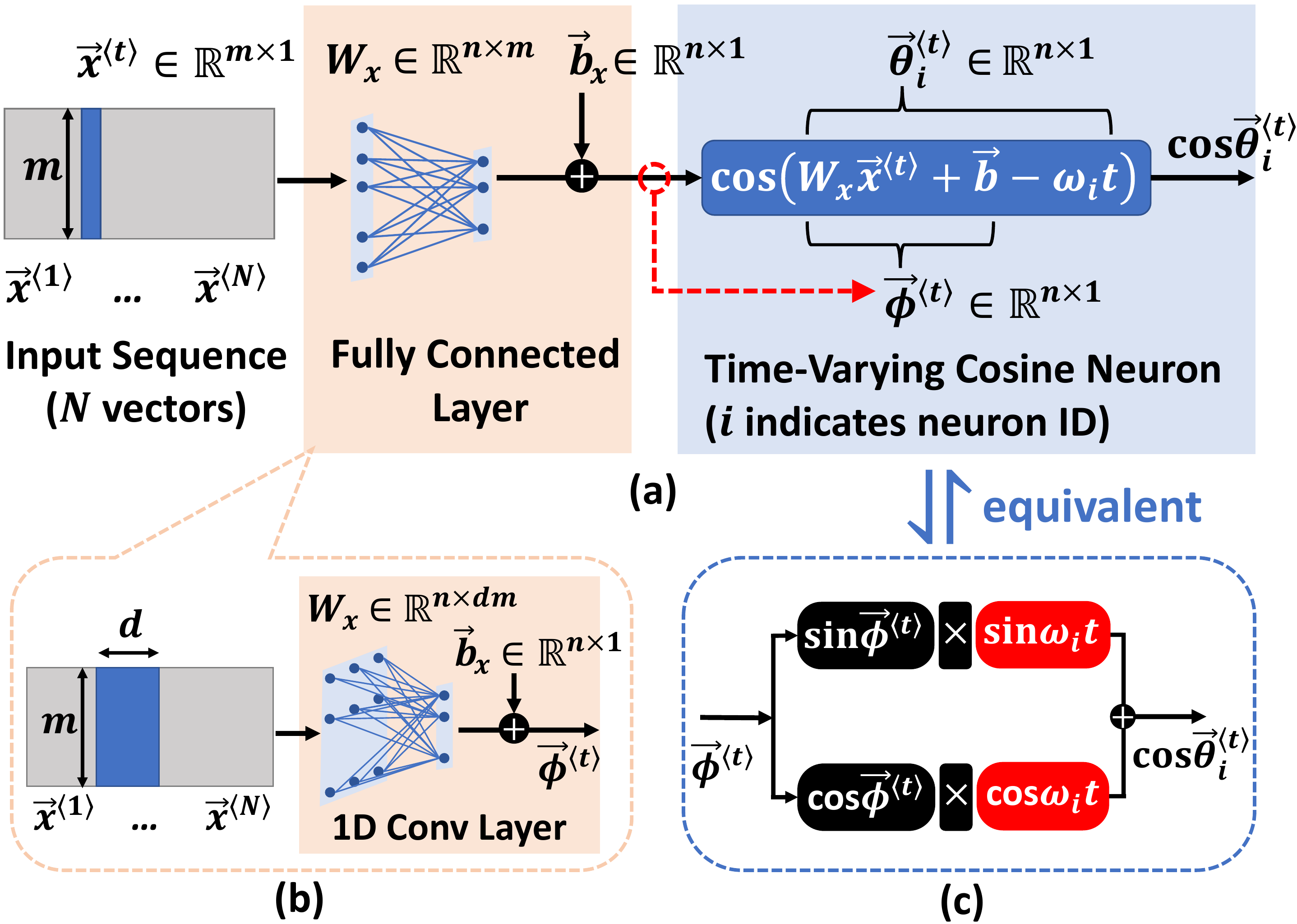}
\caption{(a) Time-Varying Cosine neuron with fully-connected input layer. (b) One dimensional convolution layer can also be used as input layer. (c) TV-Cosine neuron is mathematically equivalent to projecting sine and cosine activations respectively onto  $\sin \omega_i t$ and $\cos \omega_i t $ channels of discrete frequencies. }
\label{fig:TV-Cosine Neuron}
\end{figure}
%figure 3 merged into figure 2.
%% are figures 2 and 3 logically equivalent? Derived in the paper? -now fig.2 and 3 are merged

\begin{figure}[ht!] 
\centering
\includegraphics[width=3.3in]{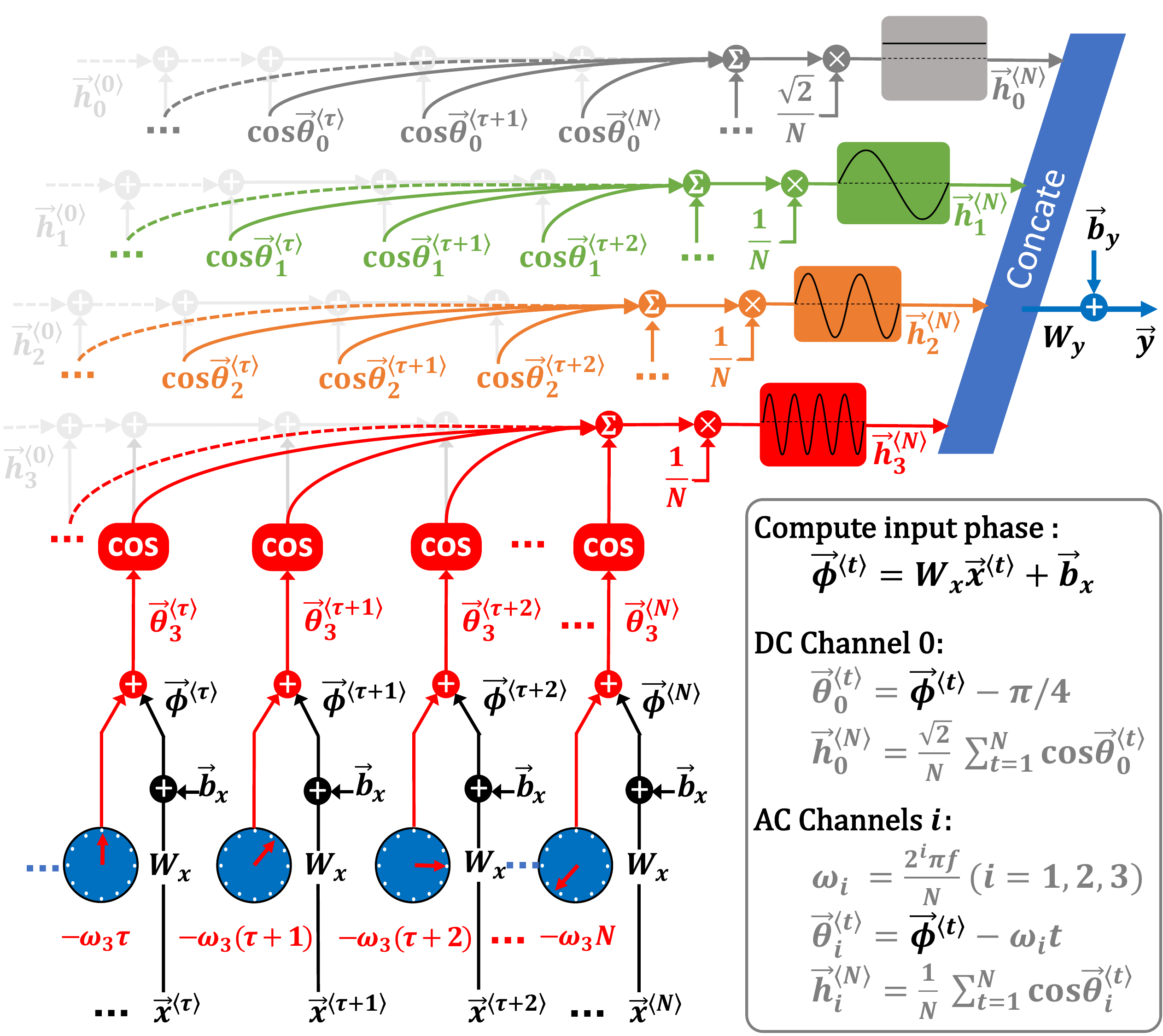}
\caption{O-FNN forward propagation (full parallelism).}
\label{fig:clock_NN_forward_dual_formats}
\end{figure}

\subsection{Time-Varying Cosine (TV-Cosine) Neuron for Sequential Processing}

We propose temporal artificial neuron named Time-Varying Cosine (TV-Cosine) neuron. As shown in Fig.\ref{fig:TV-Cosine Neuron}(a), the proposed neuron has a cosine activation, which has an input  $\theta_i^{\langle t \rangle}$ obtained from a superposition of a fully connected layer's output $\phi^{\langle t \rangle}$ and a time-varying phase modulating term $\omega_i t$. Mathematically, each TV-Cosine neuron is equivalent to a sine neuron and a cosine neuron respectively, projected to a sine and a cosine oscillating functions, as shown in Fig.\ref{fig:TV-Cosine Neuron}(c). Each TV-Cosine neuron has a specific oscillating frequency $\omega_i$, where the subscript $i$ indicates the frequency channel. The frequencies for different channels {$\omega_i$} are hyper parameters of the proposed model. In sequential processing tasks, the fully connected input layer in Fig.\ref{fig:TV-Cosine Neuron}(a) can also be replaced with one dimensional convolutional layer as shown in Fig.\ref{fig:TV-Cosine Neuron}(b), in which, $d$ consecutive input vectors are fed to the TV-Cosine neuron at each time-step. We observed that using 1D convolutional layer with a small slicing window ($d=3$) and small stride ($s=1$) slightly improves the accuracy, at the cost of slightly increasing the number of weight parameters.
% Intuitively, TV-Cosine neurons in O-FNN perform a special form of discrete Fourier transform by projecting the whole input sequence to a spectrum of various frequencies. 

\subsection{Forward pass in O-FNN}
%% Can we modify the subsection title? Forward pass in O-FNN? - done

We propose O-FNN architecture consisting of TV-Cosine neurons for efficient learning of sequential tasks. A simple example of O-FNN containing four TV-Cosine neurons with distinctive channel frequencies is shown in Fig.\ref{fig:clock_NN_forward_dual_formats}. The forward flow of O-FNN is unrolled in time as shown along the horizontal direction. 
%The number of channels equal the number of TV-Cosine neurons. 
Each channel traces and accumulates outputs of one TV-Cosine neuron across all time-steps. The final hidden states of all neurons are concatenated before further processing by a fully connected read-out layer. As discussed earlier, each TV-Cosine neuron equivalently projects activations of sine and cosine activations onto oscillating sine and cosine channels with various frequencies. Hence, the forward flow of O-FNN in Fig.\ref{fig:clock_NN_forward_dual_formats} is mathematically equivalent to a special form of Discrete Fourier Transform (DFT). 

%As shown in Fig.\ref{fig:clock_NN_forward_dual_formats}(b), the frequency channels are color matched between the two forms of computation. The sequential inputs are fed to the sine and cosine activations through a fully connected layer. Subsequently, the sine neuron's outputs are fed to block A to perform a Discrete Sine Transform (DST), and the cosine neuron's outputs are fed to block B to perform a Discrete Cosine Transform (DCT). The summation of transformation results $\vec{h}_{i (A)}^{\langle N \rangle}$ and $\vec{h}_{i (B)}^{\langle N \rangle}$ of the two blocks equals the final hidden state $\vec{h}_{i}^{\langle N \rangle}$ accumulated along the corresponding channel in Fig.\ref{fig:clock_NN_forward_dual_formats}(a), which can be concatenated and further processed by the fully connected read-out layer.

TV-Cosine neurons in O-FNN can be categorized into one DC neuron and a few AC neurons depending on the neuron input. For example, at each time-step $t$, an input phase $\vec{\phi}^{\langle t \rangle}$ is computed as shown by Eq.\ref{eq1}. Input to DC neuron equals $\vec{\phi}^{\langle t \rangle}$ subtracts a constant value $\frac{1}{4} \pi$ as shown by Eq.\ref{eq2}. Inputs to AC neurons equal $\vec{\phi}^{\langle t \rangle}$ subtracts a time-driven phase modulating term $\omega_i t$ as shown by Eq.\ref{eq3}. The angular velocity $\omega_i$ of different AC neurons are computed by Eq.\ref{eq4}, in which $f$ is a hyper-parameter used to determine the running speed of the clock in each AC neuron.

\begin{equation} \label{eq1}
\vec{\phi}^{\langle t \rangle} = W_x \vec{x}^{\langle t \rangle} + \vec{b}_x \quad \left(t=1,...,N\right)
\end{equation}

\begin{equation} \label{eq2}
\vec{\theta}_0^{\langle t \rangle} = \vec{\phi}^{\langle t \rangle} - \frac{1}{4} \pi \quad \left(t=1,...,N\right)
\end{equation}

\begin{equation} \label{eq3}
\vec{\theta}_i^{\langle t \rangle} = \vec{\phi}^{\langle t \rangle} - \omega_i t \quad \left(t=1,...,N\right)
\end{equation}

\begin{equation} \label{eq4}
\omega_i = \frac{2^i \pi f}{N} \quad \left(i=1,2,3\right)
\end{equation}

The forward flow of O-FNN is a fully parallelizable architecture as shown in Fig.\ref{fig:clock_NN_forward_dual_formats}. Concretely, neuronal activation of all time steps can be computed in parallel if the whole input sequence is already known. The process of forward propagation can be explained as follows. First, all hidden states are initialized to zero. Next, input at each time step can be fed into the parallelizable architecture, and the cosine activation from each time step can be computed following Eq.\ref{eq1} - \ref{eq4}. Subsequently, the final states of DC and AC neurons in O-FNN are computed by summing and averaging of cosine neurons' outputs over all time steps, following Eq.\ref{eq5}(a) and Eq.\ref{eq6}(a). The summation for the AC-channel neurons outputs is equivalent to combining a sine neuron's DST and cosine neuron's DCT terms.  Finally, the resulting hidden coefficients are concatenated and fed to the final read-out layer as shown by Eqs.\ref{eq7} and \ref{eq8}. Note that, by subtracting a constant phase $\frac{1}{4} \pi$ in Eq.\ref{eq2} (and multiplying a coefficient of $\sqrt{2}$), we manage to obtain a unified formal expression of hidden states of both DC and AC channels. 

\begin{subequations} \label{eq5}
\begin{alignat}{4}
\vec{h}_0^{\langle N \rangle} &=  \frac{\sqrt{2}}{N} \left( \vec{h}_0^{\langle 0 \rangle} + \sum_{t=1}^{N} \cos \vec{\theta}_0^{\langle t \rangle} \right) \\
&= \frac{1}{N} \sum_{t=1}^{N} \left( \sin \vec{\phi}^{\langle t \rangle} + \cos \vec{\phi}^{\langle t \rangle} \right)
\end{alignat}
\end{subequations}

\begin{subequations} \label{eq6}
\begin{alignat}{4}
\vec{h}_i^{\langle N \rangle} &= \frac{1}{N} \left( \vec{h}_i^{\langle 0 \rangle} + \sum_{t=1}^{N} \cos \vec{\theta}_i^{\langle t \rangle} \right) \\
&= \medmath{ \frac{1}{N} \sum_{t=1}^{N} \left[ \sin \vec{\phi}^{\langle t \rangle} \sin (\omega_i t) + \cos \vec{\phi}^{\langle t \rangle} \cos (\omega_i t) \right] }
\end{alignat}
\quad \quad \quad \quad $ \left( \mbox{in which} \ \ i=1,2,3 \ \ \mbox{for AC channels} \right) $
\end{subequations}

\medskip

\begin{equation} \label{eq7}
\vec{h}_{cat}^{\langle N \rangle} = \mbox{concate} \! \left[ \vec{h}_0^{\langle N \rangle}; \vec{h}_1^{\langle N \rangle}; \vec{h}_2^{\langle N \rangle}; \vec{h}_3^{\langle N \rangle} \right]
\end{equation}
\quad \quad \quad \quad \quad (concatenate along dimension 1)

\medskip

\begin{equation} \label{eq8}
\vec{y} = W_y \vec{h}_{cat}^{\langle N \rangle} + \vec{b}_y
\end{equation}

The compact model size and high computational efficiency stem from the fact that O-FNN does not require the square matrix $W_{hh}$ and the associated matrix multiplication operations $\vec{h}^{\langle t \rangle} = W_{hh} \left[ \vec{h}^{\langle t-1 \rangle} + f(W_x \vec{x}^{\langle t \rangle} + \vec{b}_x) \right] $ that occur at every time-step in regular RNNs. The superior accuracy of O-FNN can be attributed to operating in the frequency domain. The DC and low frequency AC channels efficiently capture long time dependencies from sequential data, whereas the high frequency AC channels provide short-term memory. Compared to approaches that perform spectrum analysis on raw input data or ReLU/Sigmoid neurons, our approach eliminates multiplication operations required to perform the transformation from time to frequency domain. For example, we can perform a regular DFT on neuron with activation function $f$ (ReLU or Sigmoid). However, as described by Eqs.\ref{eq9} (a) and (b), at least one multiplication is required at each time-step to complete the transformation. On the other hand, only addition operations are used in our approach to perform the special form of DFT in Fig.\ref{fig:clock_NN_forward_dual_formats} using TV-Cosine neurons and O-FNN architecture in Fig.\ref{fig:clock_NN_forward_dual_formats}. The required cosine computation can be implemented using a look-up table. The required extra memory space $O(N)$ ($N$ is sequence length) is negligible compared to the trainable parameters. 

\begin{subequations} \label{eq9}
\begin{alignat}{4}
\vec{h}_i^{\langle N \rangle} &= \medmath{ \frac{1}{N} \sum_{t=1}^{N} \left[ f \! \left( \vec{\phi}^{\langle t \rangle} \right) \sin (\omega_i t) + f \! \left( \vec{\phi}^{\langle t \rangle} \right) \cos (\omega_i t) \right] } \\
&= \frac{\sqrt{2}}{N} \sum_{t=1}^{N} \left[ f \! \left( \vec{\phi}^{\langle t \rangle} \right) \cos \left( \omega_i t - \frac{\pi}{4} \right) \right]
\end{alignat}
\quad \quad \quad \quad $ \left( \mbox{in which} \ \ i=1,2,3 \ \ \mbox{for AC channels} \right) $
\end{subequations}

\medskip
\medskip

The number of AC neurons (channels) used in O-FNN is a hyper-parameter. According to our simulation results, using only two to three AC neurons are sufficient to obtain the SOTA accuracies across various datasets we tested. Increasing the number of AC neurons beyond four causes accuracy degradation due to over-fitting. As can be seen in the result section, O-FNN requires far fewer neurons and smaller memory footprint while achieving superior accuracy than today's SOTA models.

\begin{figure}[t!]
\centering
\includegraphics[width=3.3in]{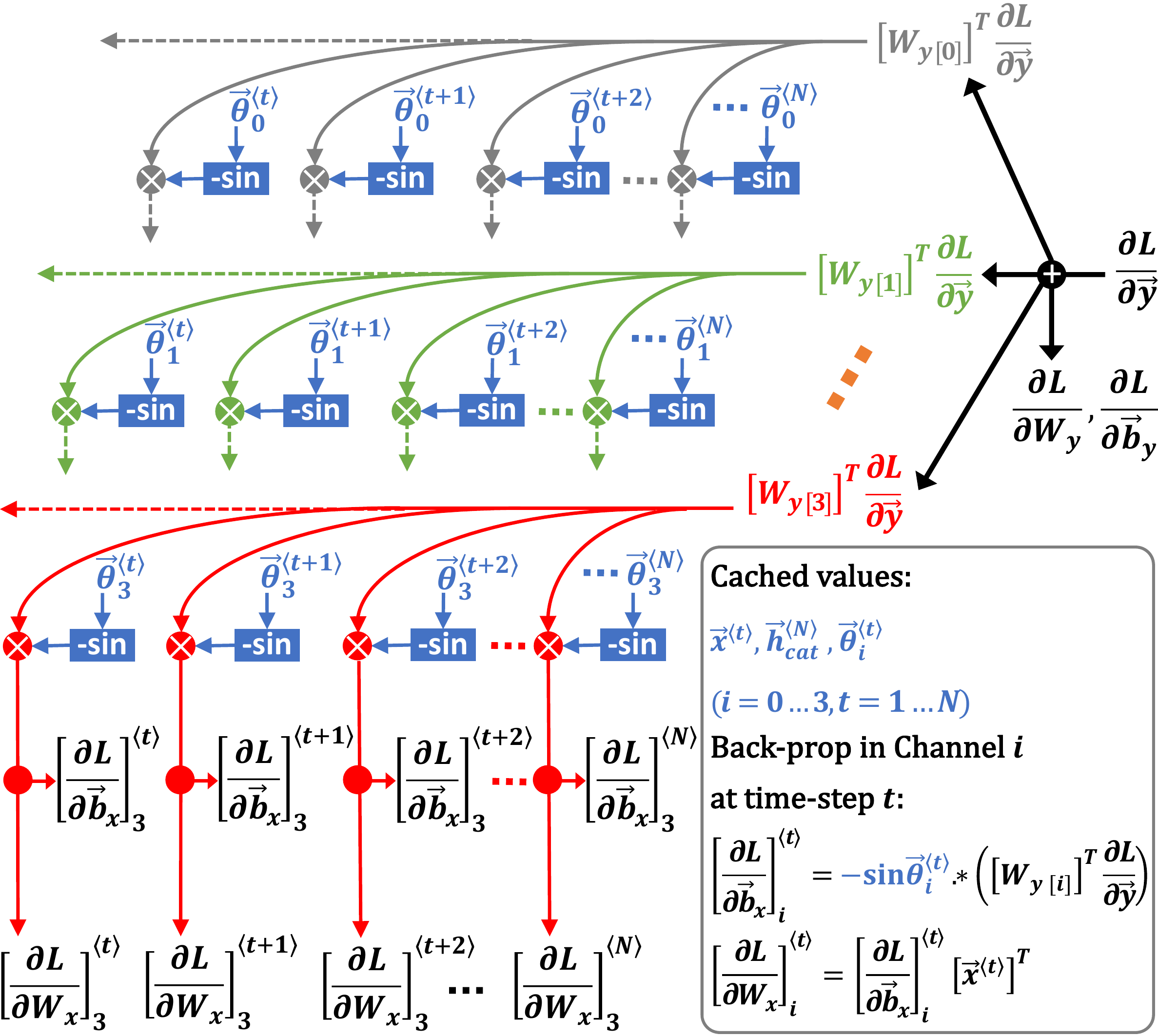}
\caption{O-FNN backward propagation (full parallelism).}
\label{fig:clock_NN_backward_cosine_format}
\end{figure}

%\subsection{O-FNN Backward Flow}
\subsection{Backward pass in O-FNN}
The back-propagation of O-FNN is shown in Fig.\ref{fig:clock_NN_backward_cosine_format}. During forward propagation, the input sequence $\vec{x}^{\langle 1 \rangle},...,\vec{x}^{\langle N \rangle}$, the concatenated final hidden states $\vec{h}_{cat}^{\langle N \rangle} \in \mathbb{R}^{4n \times 1}$, and the inputs $\vec{\theta}_i^{\langle t \rangle}$ to all TV-Cosine neurons at all time-steps have been cached and will be used in back-propagation. Gradient w.r.t. output $\frac{\partial L}{\partial \vec{y}} \in \mathbb{R}^{d \times 1}$ first propagates through the fully connected read-out layer and the parameters in this layer can be updated using Formulas \ref{eq10} and \ref{eq11}, where $\eta$ is learning rate. Next as shown in Formula \ref{eq12}, the weight matrix $W_y \in \mathbb{R}^{d \times 4n}$ of the read-out layer is decomposed into four sub-matrices $W_{y \left[i \right]} \in \mathbb{R}^{d \times n}$, where $(i=0,...,3)$. Gradients w.r.t each one of the backward channels can be computed as $\left[ W_{y \left[ i \right] } \right]^T \! \frac{\partial L}{\partial \vec{y}}$ as shown in Fig.\ref{fig:clock_NN_backward_cosine_format}. No BPTT is used in O-FNN because all gradients w.r.t. input bias and weights at all time-steps and channels can be computed in a fully parallel fashion as shown by Eqs.\ref{eq13} and \ref{eq14}, respectively. Finally, the parameters of input layer are updated using the averaged gradients across all time-steps and channels as indicated by \ref{eq15} and \ref{eq16}.

\begin{equation} \label{eq10}
\vec{b}_y \leftarrow \vec{b}_y + \eta \frac{\partial L}{\partial \vec{y} }
\end{equation}

%\medskip

\begin{equation} \label{eq11}
W_{y} \leftarrow W_{y} + \eta \frac{\partial L}{\partial \vec{y}} \left[ \vec{h}_{cat}^{\langle N \rangle} \right]^T
\end{equation}

%\medskip

\begin{equation} \label{eq12}
\mbox{concate} \! \left[ W_{y \left[ 0 \right]}; W_{y \left[ 1 \right]}; W_{y \left[2 \right]}; W_{y \left[ 3 \right]} \right] \leftarrow W_y
\end{equation}
\quad \quad \quad \quad \quad (concatenate along dimension 1)

\medskip

\begin{equation} \label{eq13}
\left[ \frac{\partial L}{\partial \vec{b}_{x}} \right]_i^{\langle t \rangle} =-\sin{\vec{\theta}_i^{\langle t \rangle}} \! .\!* \! \left( \left[ W_{y \left[ i \right]} \right]^T \frac{\partial L}{\partial \vec{y}} \right)
\end{equation}
\quad \quad \quad $ ( \mbox{in which} \ .\!* \ \mbox{is element-wise multiplication})$ 

\medskip

\begin{equation} \label{eq14}
\left[ \frac{\partial L}{\partial W_{x}} \right]_i^{\langle t \rangle} = \left[ \frac{\partial L}{\partial \vec{b}_{x}} \right]_i^{\langle t \rangle} \left[ \vec{x}^{\langle t \rangle} \right]^T
\end{equation}

%\medskip

\begin{equation} \label{eq15}
\vec{b}_{x} \leftarrow \vec{b}_{x} + \frac{\eta}{4N} \sum_{i=0}^3 \sum_{t=1}^{N} \left[ \frac{\partial L}{\partial \vec{b}_{x}} \right]_i^{\langle t \rangle}
\end{equation}

%\medskip

\begin{equation} \label{eq16}
W_{x} \leftarrow W_{x} + \frac{\eta}{4N} \sum_{i=0}^3 \sum_{t=1}^{N} \left[ \frac{\partial L}{\partial W_{x}} \right]_i^{\langle t \rangle}
\end{equation}

\medskip

By eliminating hidden state weight matrix $W_{hh}$ and the time consuming BPTT in the backward pass, training O-FNN can be faster with considerable improvement in energy efficiency. Moreover, the occurrence of exploding and vanishing gradients due to multiplicative errors across numerous time steps is also prevented in absence of BPTT. Unlike GRUs/LSTMs, which employ extra states and gates to retain memory for long-term dependencies, O-FNN resorts to transforming the input sequences to the frequency domain, and focuses on training the network based on the spectral information. Since O-FNN is designed to accumulate equal contribution from all time-steps, the updates of parameters is determined by the averaged gradients, leading to an equivalent "mini-batch" effect during training. Such equivalent "mini-batch gradient descent" based training in O-FNN might be able to explain the fast convergence with O-FNN compared to other SOTA sequential models, as is demonstrated in the Results and Discussion Section. 

% As is discussed in further details in the Results and Discussion Section, O-FNN demonstrate benefits of having significantly smaller models and faster convergence at training, when compared with most SOTA RNN models. 

%% Confusing! I thought without the matrix its just like regular RNN -done: rewritten

\section{Results and Discussion}
We will discuss the tuning of hyper-parameters for optimizing the network architecture. Specifically we investigate the impact of the base frequency $f$ and the total number of channels ($\#chs$) on the model's learning capability. Then we will present the results of applying the proposed O-FNN to a diverse group of learning tasks. 
We conducted experiments on an NVIDIA GeForce GTX 1080 GPU. We perform hyper-parameters tuning based on a grid search. The best performing results is based on the highest mean values of validation accuracy averaged over 10 random initialization of trainable parameters. %, where the results of the best performing O-FNN (based on the validation set) are reported. 
We used an exponential decaying learning with an initial learning rate 1e-3 for all the experiments. For the IMDB task, an exponential decaying factor 0.3 is used, and 0.7 is used for other tasks.

\begin{figure*}[ht!]
\centering
\includegraphics[width=6.8 in]{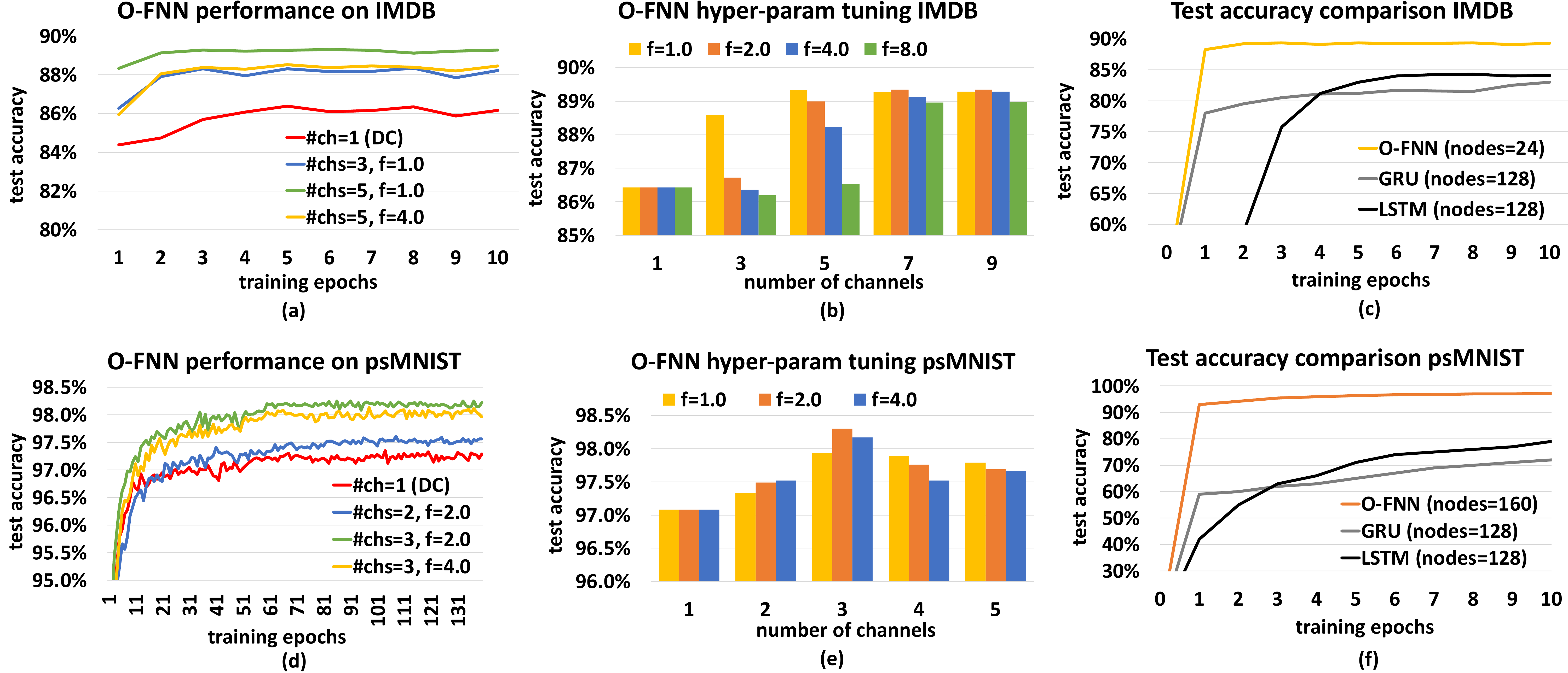}
\caption{(a) O-FNN performance vs training epochs on IMDB dataset (b) Hyper-parameters tuning on IMDB dataset (c) Test accuracy comparison on IMDB dataset (d) O-FNN performance vs training epochs on psMNIST dataset (e) Hyper-parameters tuning on psMNIST dataset (f) Test accuracy comparison on psMNIST dataset.}
\label{fig:results_tuning}
\end{figure*}

\subsection{Hyper-parameters Tuning and Understanding}
We start with optimizing hyper-parameters for the IMDB sentiment classification task. 
IMDB dataset contains 50K movie reviews from IMDB users. The sentiments of the reviews written by IMDB users are labeled as either positive ("1") or negative ("0") \cite{Maas2011Learning}. The binary sentiment classification task is to differentiate a user review being positive or negative. The 50K dataset is split evenly for training and testing (25K for each). %The IMDB data set  is a collection of 50k movie reviews, where 25k reviews are used for training (with 7.5k of these reviews used for validating) and 25k reviews are used for testing. The task is to decide whether a movie review is positive or negative. We follow the standard procedure by initializing the word embedding with pretrained 50d GloVe \cite{Brochier2019Global} vectors and restrict the dictionary to 25k words. 
As for word embedding, we apply the pretrained 50d Glove to vectorize the vocabulary \cite{Brochier2019Global}.
As is shown in Fig.\ref{fig:results_tuning}(a), adding AC channels of various base frequency $f$ improves the learning performance compared to the case of using DC channel only, corroborating the importance of capturing AC features for learning the long sequences of texts. For IMDB dataset, it is found that base frequency of $f$ around 1.0 works well. Increasing $f$ above 2.0 hurts the performance, possibly due to the loss of low-frequency information which is critical for learning long-time dependencies. It is also observed that increasing the total number of channels from 3 to 5 further improve the test accuracy with a more stabilized learning curve versus number of epochs. Fig.\ref{fig:results_tuning}(b) summarizes the results of hyper-parameter tuning for IMDB dataset. The high efficiency of the Time-Varying Cosine neuron is clearly demonstrated, as only 3-5 channels are sufficient to extract a rich set of features in the time-domain. Adding more channels (above 5) shows no benefit while increasing the model size.

In addition to text classifications, we also evaluate the tuning of hyper-parameters for the task of image classification. We look into digit classification on permuted sequential MNIST dataset (psMNIST). The psMNIST is a modification of sequential MNIST \cite{Le2015ASimple} (sMNIST), where a fixed permutation is applied to the stream of individual pixels based on MNIST digits \cite{Lecun1998Gradient}.
%In permuted sequential MNIST (psMNIST) a fixed random permutation is applied to the Sequential MNIST (sMNIST) dataset, making it more challenging for RNNs to learn the pattern in an MNIST \cite{Lecun1998Gradient} digit. 
The psMNIST provides sequences of pixels with a sequence length T = 784, which is a few times longer than the sequence length of IMDB reviews (averaged at about 250 words). As is shown in Fig.\ref{fig:results_tuning}(d), we observe improvement of performance as AC channels are added into the network, similarly to the observation from the IMDB experiment.
Moreover, it is found that higher base frequency is desirable to cope with the increased sequence length in psMNIST. As is further discussed in the following, the choice of optimized base frequency is dependent on the characteristics of dataset such as the sequence length and the ratio of pattern length to total sequence length.
As is summarized in Fig.\ref{fig:results_tuning}(e), the best performance occurs when number of channels equals 3 with the base AC frequency $f$=2.0.
The observed degradation of test accuracy for number of channels higher than 3 is due to overfitting, as a large number of neurons are connected to process single-pixel sequences. 
%Performance degradation was observed as we increase the number of channels beyond five. Due to the fact that we use more units per channel in this experiment than IMDB, increasing the number of channels further increases total number of neurons which causes the model to overfit.
\subsection{Results on Various Datasets}
We apply the O-FNN to various sequential learning tasks, with hyper-parameters optimized for each dataset and task. In addition to IMDB and permuted sequential MNIST as discussed above, we further apply the O-FNN model to process noise padded CIFAR-10 \cite{chang2019antisymmetricrnn} and human activity recognition (HAR-2) datasets \cite{Anguita2012Human}\cite{Kusupati2018FastGRNN}.
%Sensor data of 6 human activities tracked by accelerometers and gyroscopes from Samsung Galaxy smartphones \cite{Anguita2012Human}
The results of O-FNN on the four datasets are summarized in Table \ref{tb:all}. For each task, we compare the O-FNN with a few baseline models such as various types of LSTM (\cite{Pandey2020AComparative}) and GRU (\cite{Dey2017Gatevariants}) as well as some recent results from the literature (such as coRNN\cite{rusch2021coupled}) for comparison.
%First, we compare the proposed O-FNN model with GRU and LSTM baselines on both IMDB sentiment analysis and psMNIST digit classification. 
As is shown in Fig.\ref{fig:results_tuning}(c) and Fig.\ref{fig:results_tuning}(f), the proposed O-FNN reaches the highest test accuracy with faster convergence on both IMDB and sMNIST benchmark experiments. 
% Table 1 summarizes the results for O-FNN and other recently published models including LSTM \cite{Pandey2020AComparative}, GRU \cite{Dey2017Gatevariants}, \hl{and CoRNN} \cite{rusch2021coupled}. \hl{The listed results for LSTM and GRU are from} \cite{rusch2021coupled}, \hl{where models with a smaller number (128) of hidden units are trained}. % which are trained similarly and have less number of hidden units, i.e. 128
For IMDB, we focus the comparison with models that have a relatively small number (128) of hidden units. We observe that O-FNN achieves significantly higher accuracy with 32$\times$ fewer neurons, 35$\times$ fewer parameters and being 10$\times$ faster in training. As for the psMNIST benchmark, our proposed model reaches higher than 93$\%$ test accuracy within 3 epochs, outperforming both LSTM \cite{vanderwesthuizen2018unreasonable} \cite{Helfrich2018Orthogonal} and GRU \cite{Hafner2017Learning} \cite{Chang2017Dilated}. 
%\hl{As is summarized in Table 2, we compare the test accuracy for O-FNN with RNN baselines and some recently published results on sMNIST and psMNIST}. 
The proposed O-FNN reaches the state of the art performance at sMNIST, while outperforming all single-layer RNN architectures at processing the challenging permuted sMNIST.

\begin{table*}[!t]
\centering
\caption{O-FNN in comparison with references on multiple benchmark data sets. (Numbers marked with $\ast$ and $\dagger$ are obtained from \cite{rusch2021coupled} and \cite{Kag2020RNNs} respectively).}
\label{tb:all}
\resizebox{0.85\textwidth}{!}{%
\begin{tabular}{|l|c|c|c|c|c|}
\hline
   \textbf{Task}& \textbf{Model} & \textbf{Test accuracy($\%$)} & \multicolumn{1}{l|}{\textbf{$\#$ Units}} & \multicolumn{1}{l|}{\textbf{$\#$ Params}} & \textbf{$\#$ Epochs} \\ \hline

\multirow{4}{*}{IMDB sentiment analysis} 
 %& LSTM \cite{Arjovsky2016Unitary} & 97.3/92.7 & 128 & 68k & - \\ %\hline
 & LSTM \cite{campos2018skip} & 86.8 & 128 & 220k\textsuperscript{{$\ast$}}  & - \\ 
 & GRU \cite{Dey2017Gatevariants}  & 84.8 & 128 & 99k & 100 \\ %\cline{3-6}
 & ReLUGRU \cite{Dey2017Gatevariants} & 85.2 & 128 & 99k& 100 \\
 & coRNN \cite{rusch2021coupled} & 87.4 & 128 & 46k & 100 \\ \cline{2-6}
 & \textbf{O-FNN} \textbf{[This Work]} & \textbf{89.4} & \textbf{24} & \textbf{6k} & \textbf{5} \\ %\cline{3-6}
 & \textbf{compact O-FNN} \textbf{[This Work]} & \textbf{88.6} & \textbf{4} & \textbf{1k} & \textbf{5} \\ %\cline{3-6}
\hline
\multirow{4}{*}{sMNIST / permuted sMNIST} 
 %& LSTM \cite{Arjovsky2016Unitary} & 97.3/92.7 & 128 & 68k & - \\ %\hline
 & LSTM \cite{Helfrich2018Orthogonal} & 98.9 / 92.9 & 256 & 270k & - \\ 
 & GRU \cite{Chang2017Dilated}  & 99.1 / 94.1 & 256 & 200k & - \\ %\cline{3-6}
 & Dilated GRU \cite{Chang2017Dilated} & 99.2 / 94.6 & 256 & 20k & - \\ %\cline{3-6}
 & coRNN \cite{rusch2021coupled} & 99.4 / 97.3  & 256 & 134k & 100 \\ \cline{2-6}
 & \textbf{O-FNN} \textbf{[This Work]} & \textbf{99.3 / 98.3} & \textbf{48 / 160} & \textbf{11k / 29k} & \textbf{54 / 86} \\ %\cline{3-6}

\hline
\multirow{4}{*}{Noise padded CIFAR-10} & LSTM \cite{Kag2020RNNs} & 11.6 & 128 & 64k & - \\ %\hline
 & Incremental RNN \cite{Kag2020RNNs}  & 54.5 & 128 & 11.5k & - \\ %\cline{3-6}
 & Lipshitz RNN \cite{Erichson2020Lipschitz} & 59 & 256 & 134k & 100 \\ %\cline{3-6}
 & coRNN \cite{rusch2021coupled} & 59 & 128 & 46k & 120 \\ \cline{2-6}
 & \textbf{O-FNN} \textbf{[This Work]} & \textbf{60.1} & \textbf{128} & \textbf{65k} & \textbf{39} \\ %\cline{3-6}
 \hline
%\hline
%   \textbf{Task}& \textbf{Model} & \textbf{test accuracy} & \multicolumn{1}{l|}{\textbf{\# units}} & \multicolumn{1}{l|}{\textbf{\# params}} & \textbf{\# epochs} \\ \hline
\multirow{4}{*}{HAR-2} 
 & LSTM \cite{Kag2020RNNs}  & 93.7 & 64 & 16k & - \\ %\cline{3-6}
& GRU \cite{Kag2020RNNs} & 93.6 & 75 & 16k & - \\ %\hline% & FastRNN \cite{Kusupati2018FastGRNN} & 94.5\% & 80 & 7k & - \\ \cline{3-6}
 & FastGRNN-LSQ \cite{Kusupati2018FastGRNN} & 95.6 & 80 & 7.5k\textsuperscript{$\dagger$} & 300 \\ %\cline{3-6}
% & anti.sym. RNN \cite{Kag2020RNNs} & 93.2\% & 120 & 8k & - \\ \cline{3-6}
%  & incremental RNN \cite{Kag2020RNNs} & 96.3\% & 64 & 4k & - \\ \cline{3-6}
 & coRNN \cite{rusch2021coupled} & 97.2 & 64 & 9k & 250 \\ \cline{2-6}
 %& tiny coRNN \cite{rusch2021coupled} & 96.5\% & 20 & 1k & - \\ \hline
 & \textbf{O-FNN} \textbf{[This Work]} & \textbf{96.3} & \textbf{64} & \textbf{3k} & \textbf{57} \\ %\cline{3-6}
 & \textbf{compact O-FNN} \textbf{[This Work]} & \textbf{94.7} & \textbf{16} & \textbf{0.8k} & \textbf{77} \\ \hline
\end{tabular}%
}
\end{table*}

\begin{table}[ht!]
\centering
\caption{Mean and standard deviation of O-FNN performance for experiments based on 10 re-trainings using random initialization of the trainable parameters.}
\label{tb:SNNs-accuracy-cifar10}
\resizebox{0.44\textwidth}{!}{%
\begin{tabular}{|l|c|c|c|}
\hline
  \textbf{Models} & \textbf{mean} & \multicolumn{1}{l|}{\textbf{standard deviation}} \\ \hline
  IMDB (24 units)  & 88.85\% &  0.49\% \\ \hline
  sMNIST (48 units) & 99.01\% & 0.15\%  \\ \hline
  psMNIST (160 units) & 97.73\% & 0.68\%  \\ \hline
  Noise pdded CIFAR-10 (128 units) & 59.22\% & 0.98\%  \\ \hline
  HAR-2 (64 units) &  95.87\% & 0.57\%  \\ \hline
\end{tabular}%
}
\end{table}

\begin{figure*}[ht!]
\centering
\includegraphics[trim=0 165 0 170, clip,width=7.0 in]{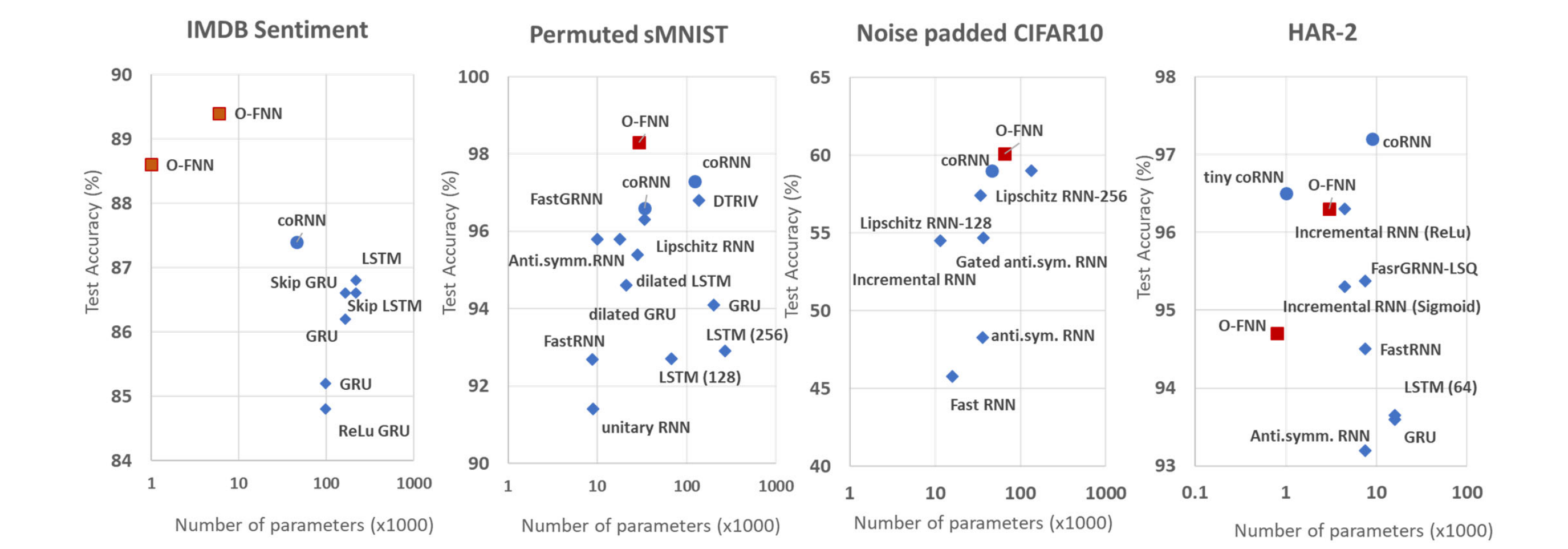}
\caption{Summary of test accuracy versus number of parameters of O-FNN in comparison with recent references. Reference points are collected from \cite{Arjovsky2016Unitary}, \cite{Dey2017Gatevariants},\cite{Chang2017Dilated} \cite{campos2018skip},\cite{Helfrich2018Orthogonal}, \cite{Kusupati2018FastGRNN},\cite{Casado2019Trivializations},\cite{chang2019antisymmetricrnn},\cite{Kag2020RNNs}, \cite{Erichson2020Lipschitz}, and \cite{rusch2021coupled}.}
\label{fig:trade-off}
\end{figure*}

The noise padded CIFAR-10 experiment poses great challenges for models to learn temporal dependence over the large number of time steps, due to the fact that the padded input sequences contain Gaussian noise in 968 out of 1000 time steps while only 32 time steps contain the serialized CIFAR10 images \cite{Krizhevsky2009Learning}. 
%the CIFAR-10 \cite{Krizhevsky2009Learning} {input images are sequentially fed into the neural network one row at each time step. Given the 3 RGB channels of an image size 32x32, the input sequence length will be 32 with the input dimension of 96. After the first 32 time steps, independent Gaussian noise is padded in the subsequent 968 time steps to form a total sequence length of 1000. In this way, the majority of input sequence is noise, making it more challenging for models to learn the temporal dependence over the large number of time steps.}
%In this test, CIFAR-10 data points \cite{Krizhevsky2009Learning} are fed to the RNN row-wise and flattened along the channels resulting in sequences of length 32. To test the long term memory, entries of uniform random numbers are added such that the resulting sequences have a length of 1000, i.e. the last 968 entries of each sequence are only noise to distract the network. 
From Table \ref{tb:all}, we observe that O-FNN outperforms other RNN architectures on this benchmark, while requiring only 65k parameters. 
% The best performance occurs when the number of channels is three (one DC channel and two AC channels). 
We found that a higher base frequency ($f$=8.0) is needed for the model to learn informative features in presence of highly concentrated input data over a small fraction of total sequence (i.e. only first 32 actual image input from 1000 total steps). The intuition is that high frequency channels are more capable of extracting localized pattern formed by input from adjacent time steps, while low frequency channels tend to get an averaged effect from input over large numbers of steps.
To this effect, it is challenging for low frequency channels to learn meaningful features since noises take a dominating proportion in those long sequences.
Our study demonstrates that tuning the hyper-parameter $f$ can provide a powerful knob for accommodating the imbalanced distribution of input information among noise padding in long sequences. 

%Furthermore, we look into the experiment on the human activity recognition data set.
% The data set is a collection of tracked human activities, which were measured by an accelerometer and gyroscope on a Samsung Galaxy S3 smartphone.
%The original 6 types of human activities monitored by sensors on Samsung Smartphones} \cite{Anguita2012Human} \hl{were grouped into two merged classes, as proposed in} \cite{Kusupati2018FastGRNN} \hl{to form the Human Activity Recognition (HAR-2) dataset. Laying, sitting, and walking upstairs form one group, while walking, standing and walking downstairs form the other group.
%Sensor data of 6 human activities tracked by accelerometers and gyroscopes from Samsung Galaxy smartphones \cite{Anguita2012Human} were binarized to obtain two merged classes (Sitting, Laying, Walking Upstairs) and (walking, standing and walking downstairs), leading to the HAR-2 data set, which was first proposed in \cite{Kusupati2018FastGRNN}. 
As for the HAR-2 dataset, we can see that O-FNN can reach close to SOTA accuracy with fewer parameters compared to most of the references. 
%We also ran this experiment on a compact O-FNN model with less than 1K parameters. 
Morevoer, we observe that even the compact version of O-FNN remain to perform at about 95\% test accuracy with less than 1K parameters. Therefore, the compact O-FNN model can be well poised for applications in edge applications and IOT devices.
%shows the result for O-FNN together with other recently published results on HAR-2. 
%We thus propose that O-FNN can be used on resource-constrained IoT micro-controllers. 
%The best performance occurs when the number of channels is three (one DC channel and two AC channels), and the base AC frequency $f$ equals 2.0.

%\vspace{1mm}
Figure \ref{fig:trade-off} summarizes the accuracy and number of parameters of the proposed O-FNN model in comparison with recent references. In plots of model test accuracy vs model size, a trade-off is typically found when the accuracy of a model increases with number of parameters. Therefore, intrinsic gain in the capability of a model will be demonstrated if the position of the model in this plot can shift up vertically. Specifically, we observe that O-FNN is overall on the top/left side of most references across all four datasets.The overall high accuracy across various tasks can be attributed to the capability of learning features at various time scales through different frequency channels.
Based on the trade-off plots, O-FNN is significantly more efficient than all the references at IMDB sentiment analysis task. As for permuted sMNIST and noise-padded CIFAR10, O-FNN outperforms the most recent references, while the separation is smaller. On HAR-2, both O-FNN and the recently proposed coRNN perform well, suggesting a good potential of utilizing neurons with oscillatory dynamics in processing temporal signals.
The compact size of O-FNN is partially attributed to the absence of the $W_{hh}$ matrix in regular RNNs, as well as the fact that our simplified Discrete Fourier Transform requires only a few frequency channels each associated with one Time-Varying Cosine neuron. The observation of fast convergence in training across various tasks can be attributed to O-FNN's fully parallelizable architecture, which makes the error back-propagation more direct than regular RNNs that requires recursive computations through time.

% %% Comment on your dataset description and results. The writeups for all the 4 cases look very similar and it feels as if you have taken two paragraphs and only substituted the results (numbers) and the corresponding names of the datasets. Is it necessary to present the results separately the way you are doing it? My suggestion would be rewrite portions.

% %% Earlier you mentioned language translation. However, there is no mention of (and results of) language translation. 
% % - yes agreed, the IMDB sentiment analysis we show is a case of text classification, many-to-one NLP task

\section{Conclusion}

A novel RNN architecture (O-FNN) based on a Time-Varying Cosine neuron model is proposed for sequential processing. Compared to conventional Fourier transform on raw input data and ReLU/sigmoidal activations, the proposed architecture provides a computationally efficient approach to extracting frequency-domain information from sequential data. 
Moreover, the O-FNN by design have full parallelism in both forward and backward passes, leading to significant simplification in training while eliminating the issue of exploding/vanishing gradients encountered by RNNs. In contrast with GRU/LSTM models, which require various types of gates and additional memory/cell states to retain long-term memory, our proposed architecture has extremely compact model size, while achieving the retention of information over long sequences through the learning of frequency-domain feature.
%% LSTMs? GRUs? 
%-done LSTM/GRU approach: add gate and long memory, gate add parameters and compute; our approach: rely on frequency domain analysis to retain long-term dependency, minimal extra parameters
We show that O-FNN is capable of handling long time dependencies with significantly smaller model size and lower computational cost, while retaining superior accuracy, faster convergence, and better error resiliency than the State-of-The-Art across various types of sequential tasks. Future avenues of work could extend the current O-FNN to address many-to-many tasks such as machine translation, or image captioning. Leveraging the potential of fast training and having compact models, O-FNN can also be further explored to process complex tasks such as video analysis.

%% In the Conclusion should we add a sentence or two towards the beginning to highlight our approach that leads to the improvement in the number of parameters, training time and accuracy.

\cleardoublepage

\cleardoublepage

{\small
\bibliography{Oscillatory_Fourier_Neural_Network}
}

\end{document}